\documentclass[letterpaper, 10 pt, conference,table,xcdraw]{ieeeconf}  

\IEEEoverridecommandlockouts                              

\usepackage{tikzpagenodes}
\usepackage{amsmath} 
\usepackage{amssymb}  
\usepackage{pdfpages}
\usepackage{pgfplotstable}
\usepackage{pgfplots}
\pgfplotsset{compat=1.17}
\usepackage{graphicx}

\usepackage{multirow}

\usepgfplotslibrary{statistics}
\pgfplotsset{
	/pgfplots/custom legend/.style={
		legend image code/.code={
			\draw [only marks,mark=square*]
			plot coordinates { 
				(0.3cm,0cm)
			};
		}, },
}

\title{\LARGE \bf
Predicting Energy Consumption and Traversal Time of Ground Robots for Outdoor Navigation on Multiple Types of Terrain
}

\author{Matthias Eder$^{1}$, Gerald Steinbauer-Wagner$^{1}$ 
\thanks{This work was partially supported by the Austrian Research Promotion Agency (FFG) with the projects RoboNav and FutureWoodTrans.}
\thanks{$^{1}$Matthias Eder and Gerald Steinbauer-Wagner are with the Institute of Software Technology, Graz University of Technology, Graz, Austria. 
	 {\tt\small \{matthias.eder, steinbauer\}@ist.tugraz.at}}%
}

\begin{document}

\maketitle
\thispagestyle{empty}
\pagestyle{empty}

\begin{abstract}
The outdoor navigation capabilities of ground robots have improved significantly  in recent years, opening up new potential applications in a variety of settings. Cost-based representations of the environment are frequently used in the path planning domain to obtain an optimized path based on various objectives, such as traversal time or energy consumption. However, obtaining such cost representations is still cumbersome, particularly in outdoor settings with diverse terrain types and slope angles. In this paper, we address this problem by using a data-driven approach to develop a cost representation for various outdoor terrain types that supports two optimization objectives, namely energy consumption and traversal time. We train a supervised machine learning model whose inputs consists of extracted environment data along a path and whose outputs are the  predicted energy consumption and traversal time. The model is based on a ResNet neural network architecture and trained using field-recorded data. The error of the proposed method on different types of terrain is within 11\% of the ground truth data. To show that it performs and generalizes better than  currently existing approaches on various types of terrain, a comparison to a baseline method is made.
\end{abstract}

\section{Introduction}
Navigation for ground robots in outdoor environments is a highly active research domain where several disciplines meet. One of these disciplines is path planning which aims to obtain an optimal path between two locations within the robot's environment based on given objectives \cite{148404}. 
Typical objectives used in path planning are the safe traversal through the environment avoiding the risk of collision \cite{9981179}, the minimization of the robot's energy consumption \cite{Wei.2022}, its traversal time \cite{6225364}, or a combination of multiple objectives.
Finding a suitable representation for objectives is still challenging, particularly for complex outdoor terrains. Simple objectives to avoid non-traversable obstacles are effective on flat surfaces on similar terrain, but they do not translate well to other types of terrain that can have a significant impact on the objective. Furthermore, depending on whether a robot goes up or down a slope in an outdoor environment, the outcome of an objective for a robot in that location may be affected. This directional influence on a given location complicates the representation of the planning space used by the planning algorithms, which is typically a cost-based representation of the environment  \cite{4650993}, also known as a cost map.
As a result of these issues, determining cost values for path planning in outdoor environments is still a challenge.
Therefore, one of the key tasks in field robotics is to define a planning space that accurately represents given objectives in order to plan an optimal path through an environment with a complex terrain \cite{Fan.2021}.
Ideally, path planning considers several objectives at the same time to find an optimal path. We take a step in this direction by using the connection of two objectives, namely energy consumption and traversal time, to create a cost representation that can be used to represent both, energy and time optimal paths.

To generate cost maps for minimizing the energy consumption of a path, some work in literature already exists using physics-based models \cite{8593845,Quann.2020} or data-driven approaches \cite{8967963,Wei.2022}. These methods, however, only take into account a limited or simplified understanding of the environment and the capabilities of the robot.  
Similar to that, there are objectives to find time-optimal paths that take into account the robot's capabilities (e.g. acceleration limits \cite{Lepetic.2003}) but do not consider how the environment affects the robot's dynamics and its limits.
However, this is a critical element in finding time-optimal paths in complex outdoor environments since specific environmental factors have an impact on the robot's capabilities and must be taken into account when planning a path through the environment.

\begin{figure}[!t]
	\centering
	\includegraphics[width=\linewidth]{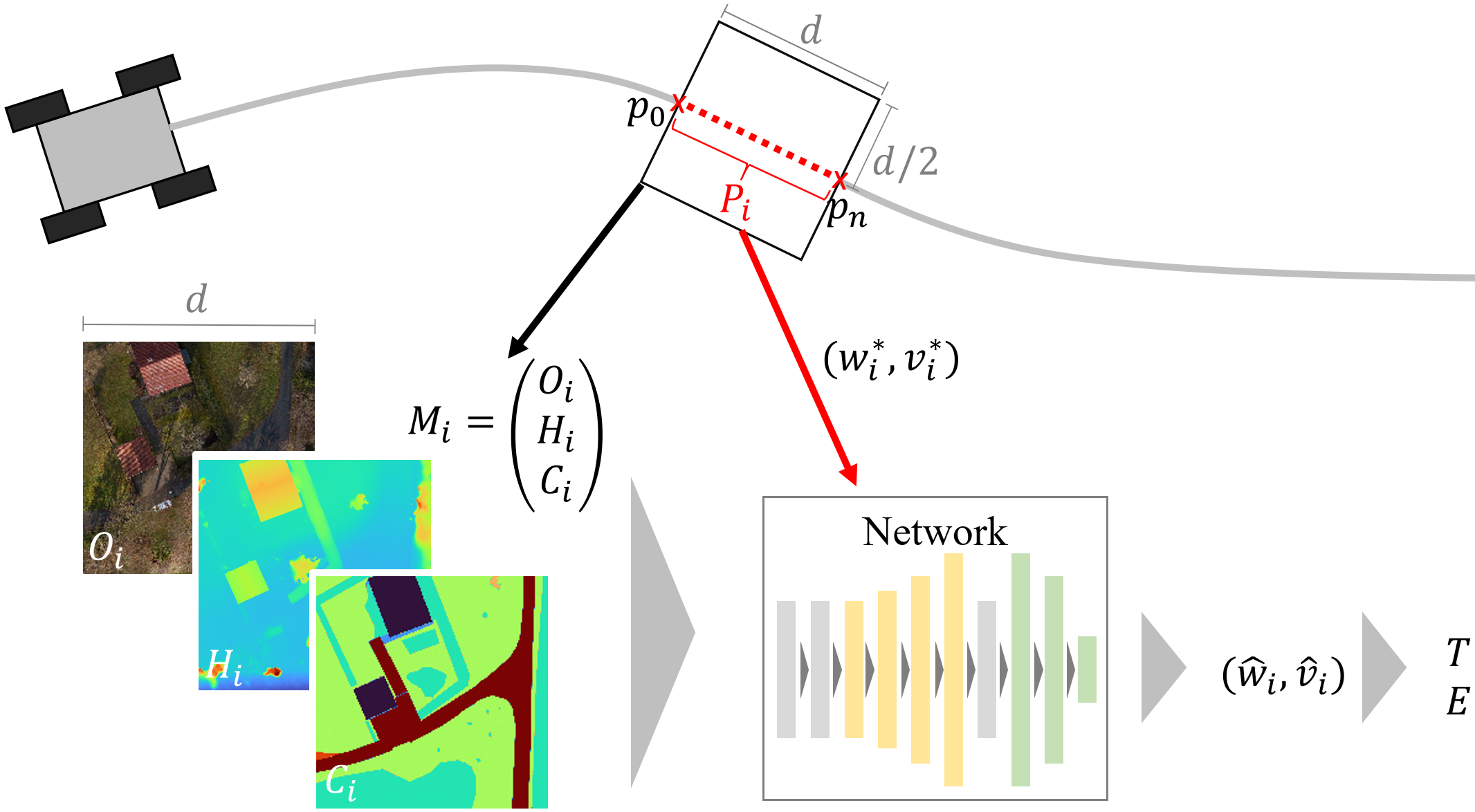}
	\caption{Overview of the proposed approach. A neural network is trained on environment information around a path segment $P_i$, which consists of the elevation $H_i$, terrain class $C_i$, and recorded orthophoto $O_i$ to predict the average power $\hat{w}_i$ and velocity $\hat{v}_i$ of the robot. The result is then used to estimate the traversal time $T$ and energy consumption $E$. }
	\label{fig:overview}
\end{figure}

To address these challenges, we present a data-driven approach that generates a cost representation for the two aforementioned path-planning objectives, energy consumption and traversal time, taking into account multiple types of outdoor terrain.
To do so, we use machine learning to train a target function that can predict the energy consumption and traversal time along a path using only environmental information along that path. 
To learn the target function, a path is divided into segments of unit length which are used to extract environment information around that path segment. 
The energy consumption and traversal time of a robot along the path are  predicted using a ResNet-based network architecture that has been trained on a regression task.
A field-deployed ground robot is used to collect data in a realistic outdoor setting on different types of terrain.
Information about the environment is extracted using photogrammetry and high-resolution aerial images. 
An overview of the proposed method is shown in Figure~\ref{fig:overview}. 

The main contributions of this paper are 1) the presentation of a machine learning-based method to predict the energy consumption and traversal time of a ground robot in outdoor terrains using environment information; 
2) the incorporation of divers environmental knowledge  to improve the prediction performance on different types of outdoor terrain; and 
3) a comparison of our method to a baseline to show that our network outperforms the baseline method and also eliminates the need to retrain the model for different types of terrain.

The remainder of this paper is structured as follows. Section~\ref{s-relatedwork} discusses related work in the field of cost estimation for outdoor traversal w.r.t. energy prediction and traversal time estimation. Section~\ref{s-problem} formulates the problem, which is learned in Section~\ref{s-method}. In Section~\ref{s-eval}, a performance evaluation of the proposed method and a comparison to a baseline approach is presented. Section~\ref{s-conclusion} concludes this paper.

\section{Related Work} \label{s-relatedwork}
Typically, a cost map representation is used as planning space for path planning algorithms to efficiently compute a path through the environment that is optimal concerning a given objective. Such a representation can either provide binary (traversable/non-traversable) \cite{Walch.2022}, continuous \cite{Zhu.2020}, or discrete cost estimates \cite{Collier.2006} following a given objective.
There are multiple objectives for optimizing the traverse in an environment such as risk of collision, smoothness, traversal time, or energy consumption \cite{Sevastopoulos.2022}. Another approach is to model costs based on a robot's motion stability by identifying unstable motion patterns that should be avoided \cite{8870912}. While some work focuses on the estimation of a single objective (e.g. \cite{Quann.2020}), others define the traversability as a combination of multiple objectives (e.g. \cite{Jian.2022}). In this work, we focus on the simultaneous estimation of the objective function for energy consumption and traversal time. 

However, the development of an energy-based cost representation is a non-trivial task for complex outdoor terrains as the energy consumption varies on different types of terrain and thus physics-based methods lack in applicability in non-uniform and rough terrains \cite{6459035,1391019,8593845}. Quann et al. aim to solve this task by acquiring friction coefficients from a designated area and terrains first and then applying Gaussian Process Regression to estimate the coefficients for new areas \cite{Quann.2020}. The method also proposes the usage of a greyscale satellite image to gain more information about unseen environments. While this method works well on terrain with available energy measurements, it is also shown that it does not generalize well in new environments where no data is available. Wei and Isler proposed a data-driven approach to predict energy consumption based on the height information around the robot using deep learning \cite{Wei.2022}. It is shown to work well on the same type of terrain but does not generalize well on other types of terrain without retraining the network on new data.
Our proposed approach aims to overcome this problem by incorporating terrain information in the trained model to accurately predict the energy consumption of a robot on multiple types of terrain.
Other factors that affect the energy consumption of a ground robot are its dynamics, e.g. acceleration and linear/angular velocity \cite{8593845}. The authors of \cite{Dogru.2019} present a method to characterize the power of a robot for a skid-steered mobile robot and optimizes the turning costs by deriving a mathematical model of friction for skid-steered platforms. Other works focus on the optimization of a trajectory for a ground robot which minimizes the energy consumed during motion \cite{Tokekar.2014}.
In this work, we focus on the estimation of the energy consumption for a given path segment which can be used as an objective to plan energy-optimal paths between two locations in the environment. 

Planning time-optimal paths through an environment is one of the main objectives in path planning and is explored in a variety of settings, primarily focusing on the robot's dynamics in a given environment \cite{Lepetic.2003,Wolek.2016}. Therefore, research has been conducted to estimate the traversability of a robotic system using different types of sensors and information, such as proprioceptive measurements \cite{9721819,Jacoff.2002}, geometry-based representations \cite{Zhuozhu.2022}, or vision sensors \cite{Broggi.2013}.  While such traversability estimates can be used in connection with the robot's theoretical dynamics to compute a time-optimal path, they do not take the influence of the environment on the robot's dynamics and its limits into account. To consider environmental factors, the authors of \cite{6225364} propose a time-optimal path planning algorithm for AUVs that integrates the continuous dynamic flow of the ocean into the planning procedure. Unlike for AUVs,  terrain rarely changes for a ground robot. However, certain environmental characteristics still have a limiting factor on the robot's capabilities and need to be considered when planning a path through the environment. In this work, we aim to improve time-optimal path planning by considering the environmental influence on the robot's dynamics. Therefore, we identify the robot's translational velocity limits in different environmental scenarios and use this information to improve the objective for time-optimal planning. 

\section{Problem Formulation} \label{s-problem}
To present the used method for predicting energy consumption and traversal time, we first formally describe the machine learning problem with the corresponding concepts and notations. 
Therefore, assume we have a ground robot that is driving along a path $\mathcal{P}$ in environment $\mathcal{M}$ and is able to measure its energy consumption $E$ and traversal time $T$. This work aims to predict $E$ and $T$ for any path in the environment using a target function $F$:
\begin{equation} \label{e-problem}
    (T,E) = F(\mathcal{M},\mathcal{P}).
\end{equation}
For this, we define path $\mathcal{P} = \{p_0, ..., p_N\}$ as a list of $N$ points $p_i = (x_i, y_i)^T$, and environment $\mathcal{M} = \{O, H, C\}$ as a set of three maps that store specific information about the environment. A map is represented as a two-dimensional array holding environment information for a dedicated point in the environment \cite{Kleiner.2007}.
The map $H$ represents the height map of an environment which provides information on the elevation of the area. $O$ is a map holding information of a gray-scale orthophoto from the environment and $C$ stores information about the type of terrain for each global location.

When a robot is navigated through an environment, it can measure its energy consumption $E$ using the battery voltage $V$  and its current $I$. Therefore, when driving along a path $\mathcal{P}$ the energy consumption between two points $p_s,p_g$ can be calculated as the integral over time $t$, between the two time instances $t_s, t_g$ of $p_s$ and $p_g$: $E = \int_{t=t_s}^{t=t_g}V(t)I(t) dt$. Since the energy of a robot is usually measured in discrete time intervals, the energy consumption between $p_s$ and $p_g$  can be estimated over $K$ measurements as  $E  \approx \sum_{i=1}^{K}V(t_i)I(t_i)\triangle t_i$,
where $t_0 = t_s$ is the time at point $p_s$ and $t_K=t_g$ the time at $p_g$. $\triangle t_i = t_i - t_{i-1}$ is the time difference between the current and last measurement. When the measurements of a robot are conducted in a constant interval such that $\triangle t_i = \triangle t_j, \forall i,j \in [1,K]$, the equation can be reformulated as
\begin{equation} \label{e-energy}
     E \approx \sum_{i=1}^{K}V(t_i)I(t_i)\triangle t_i =  \frac{T}{K} \sum_{i=1}^{K}V(t_i)I(t_i)
\end{equation}
where $T = t_K - t_0$ corresponds to the difference between $t_s$ and $t_g$ and thus to the traversal time as stated in Eq.~\ref{e-problem}. 

\section{Learning Method} \label{s-method}
The problem definition as formulated in Section~\ref{s-problem} is difficult to solve using machine learning, as the prediction of a path with arbitrary length requires a large number of training samples and faces the challenge of how to represent the paths as input.
Furthermore, the estimation of energy consumption $E$ is directly correlated with traversal time $T$, as shown in Eq~\ref{e-energy}. Due to their interdependence, it is challenging to independently learn both variables in a single  model.
To resolve these issues, we break down the formulation from Eq.~\ref{e-problem} into a more manageable problem that can be solved using machine learning, and then generalize it to the entire problem as stated above.

To do so, we divide a path $\mathcal{P}$ into path segments of unit length $d$, resulting in $l$ path segments $P_i = \{p^i_0,...,p^i_n\}, 1\leq i \leq l$ with $\sqrt{(p^i_0-p^i_n)^2}=d$. $d$ is assumed to be small enough so that it is a good approximation of the actual path length of $P_i$ such that $\sum_{j=1}^n\sqrt{(p^i_j - p^i_{j-1})^2} \approx d$.
Next, we define $M_i$ to be a patch of size $d\times d$, holding environment information $\mathcal{M}$ which was extracted along the path segment $P_i$. For a description of how $M_i$ is extracted, see Section~\ref{s-patch}.
Since $M_i$ holds the environment information along $P_i$ and $d$ is an approximation of the path length, it also indirectly holds  information about the path $P_i$.

Following Eq.~\ref{e-energy}, we define $w = 1/K \sum_{j=1}^{K}V(t_j)I(t_j)$ to be the average power used on a path segment $P_i$ with $t_0$ being the time at $p^i_0$ and $t_K$ being the time at $p^i_n$. 
Moreover, if $d$ is small enough, $T$ can be estimated as the length of the path segment $d$ divided by the average linear velocity $v$ of the robot on the path segment: $T \approx d/v$.

The assumptions made above allow us to resolve the challenges arising from Eq.~\ref{e-problem} by learning a prediction function in a single machine-learning model on two independent variables $w,v$ for a path of unit length $d$.
Therefore, we train a function $\hat{f}$ which takes an environment patch $M_i$ as input and predicts the average power $\hat{w}_i$ as well as the velocity of the robot $\hat{v}_i$ for traversing the patch:
\begin{equation}
    (\hat{w}_i, \hat{v}_i) = \hat{f}(M_i). 
\end{equation}
The function is trained by minimizing the normalized root mean squared error (NRMSE) between the prediction $(\hat{w}_i, \hat{v}_i)$ and the ground truth $(w^*_i, v^*_i)$, which was recorded by the robot:
\begin{equation} \label{e-loss}
    \min \sum_{i=1}^l \sqrt{\alpha(\hat{w}_i - w_i^*)^2 + \beta(\hat{v}_i - v_i^*)^2}
\end{equation}
with $\alpha=1/\max(w^*)$ and $\beta=1/\max(v^*)$ being the normalization factors for both variables, and $l$ being the number of data samples used for training.

Using the predictions of $\hat{f}$, the target function $F$ from Eq.\ref{e-problem} can be solved for any path $\mathcal{P}$ in $\mathcal{M}$ using the predictions $\hat{f}$ for Eq.~\ref{e-energy} over all $l$ path segments in $\mathcal{P}$
\begin{equation} \label{e-full}
    (T,E) = (\sum_{i=1}^ld/\hat{v}_i, \sum_{i=1}^l\hat{w}_i * d/\hat{v}_i)
\end{equation}


\begin{figure*}
	\centering
        \includegraphics[width=0.32\linewidth]{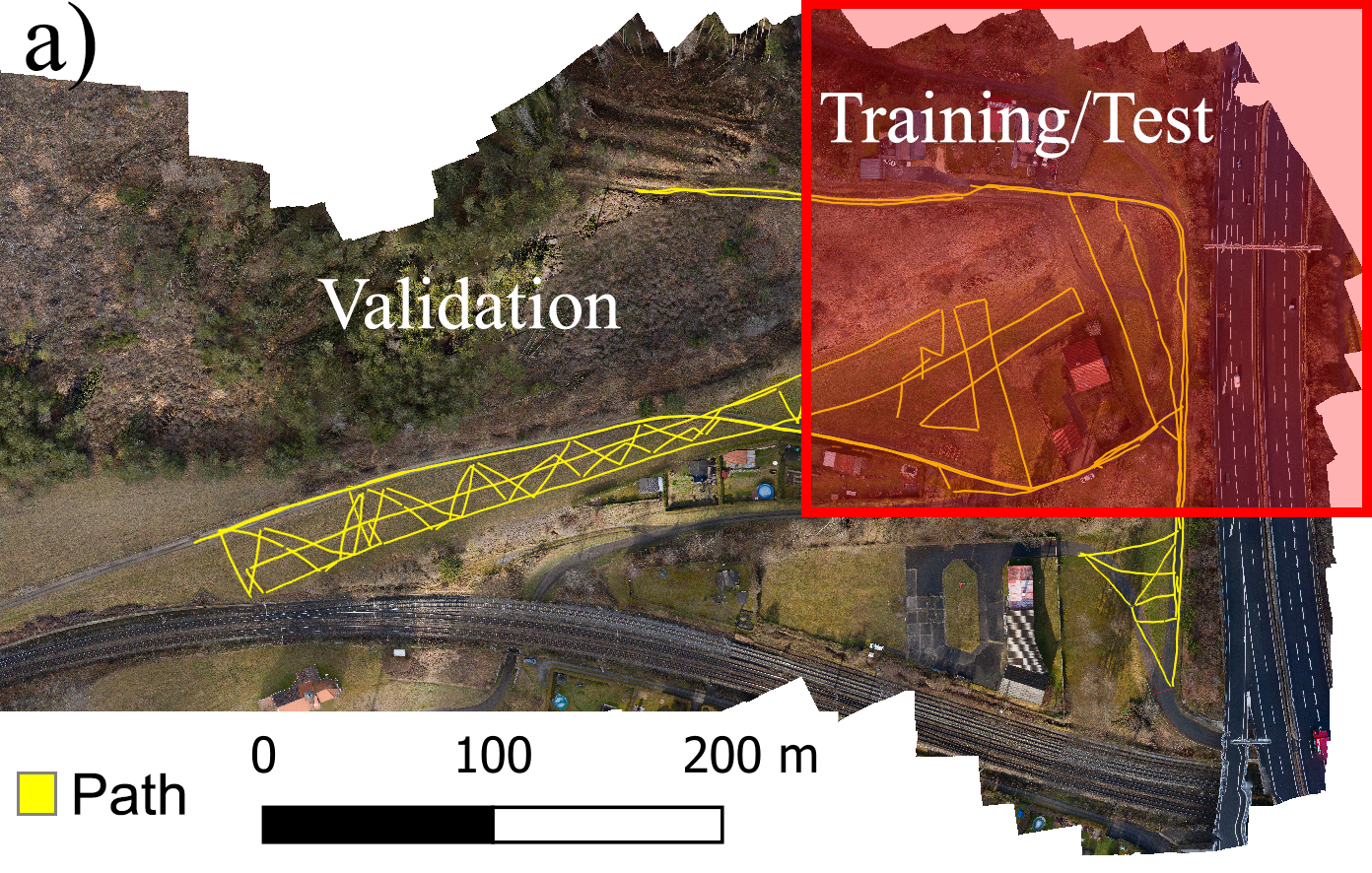}
	\includegraphics[width=0.32\linewidth]{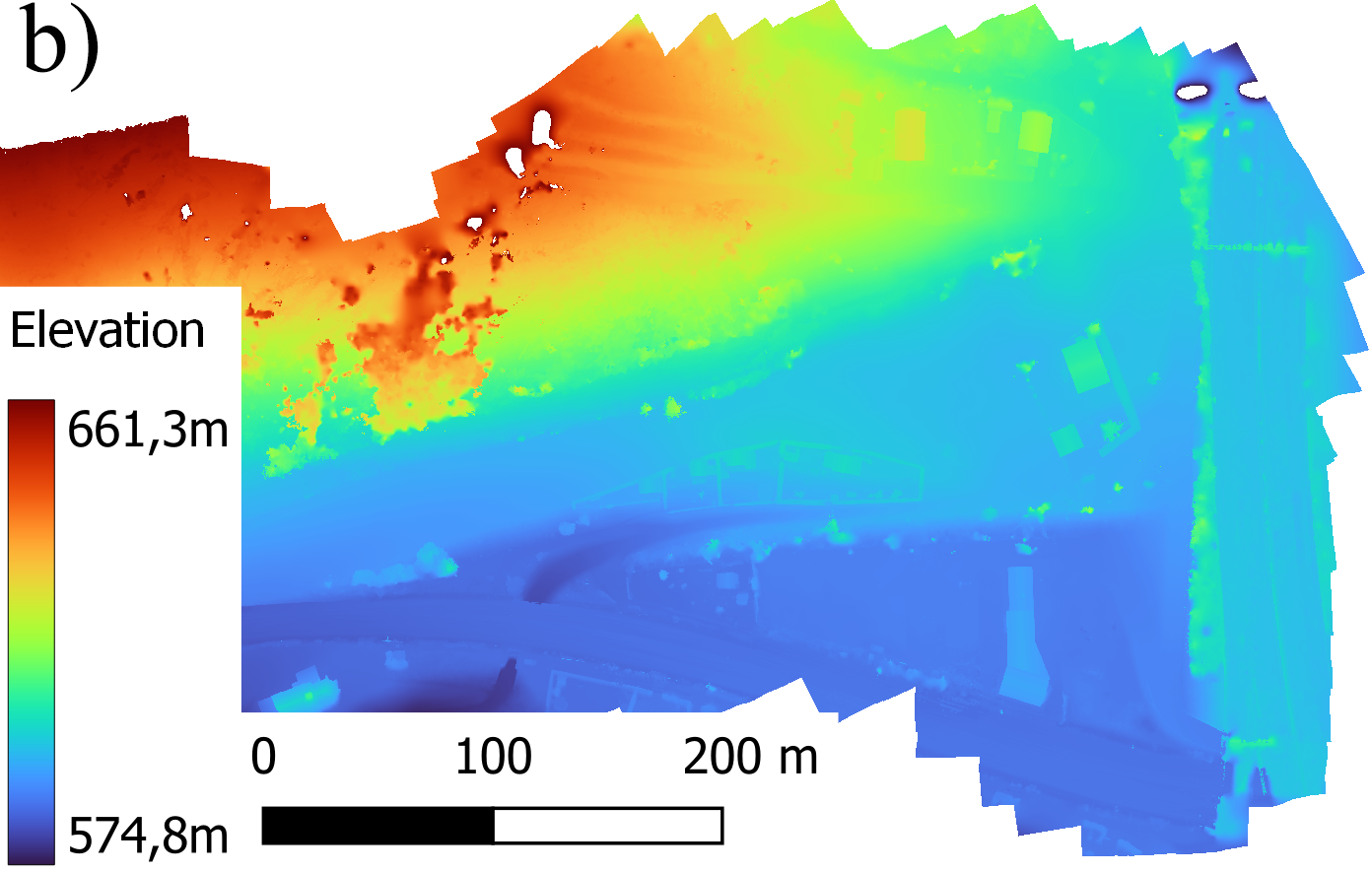}
	\includegraphics[width=0.32\linewidth]{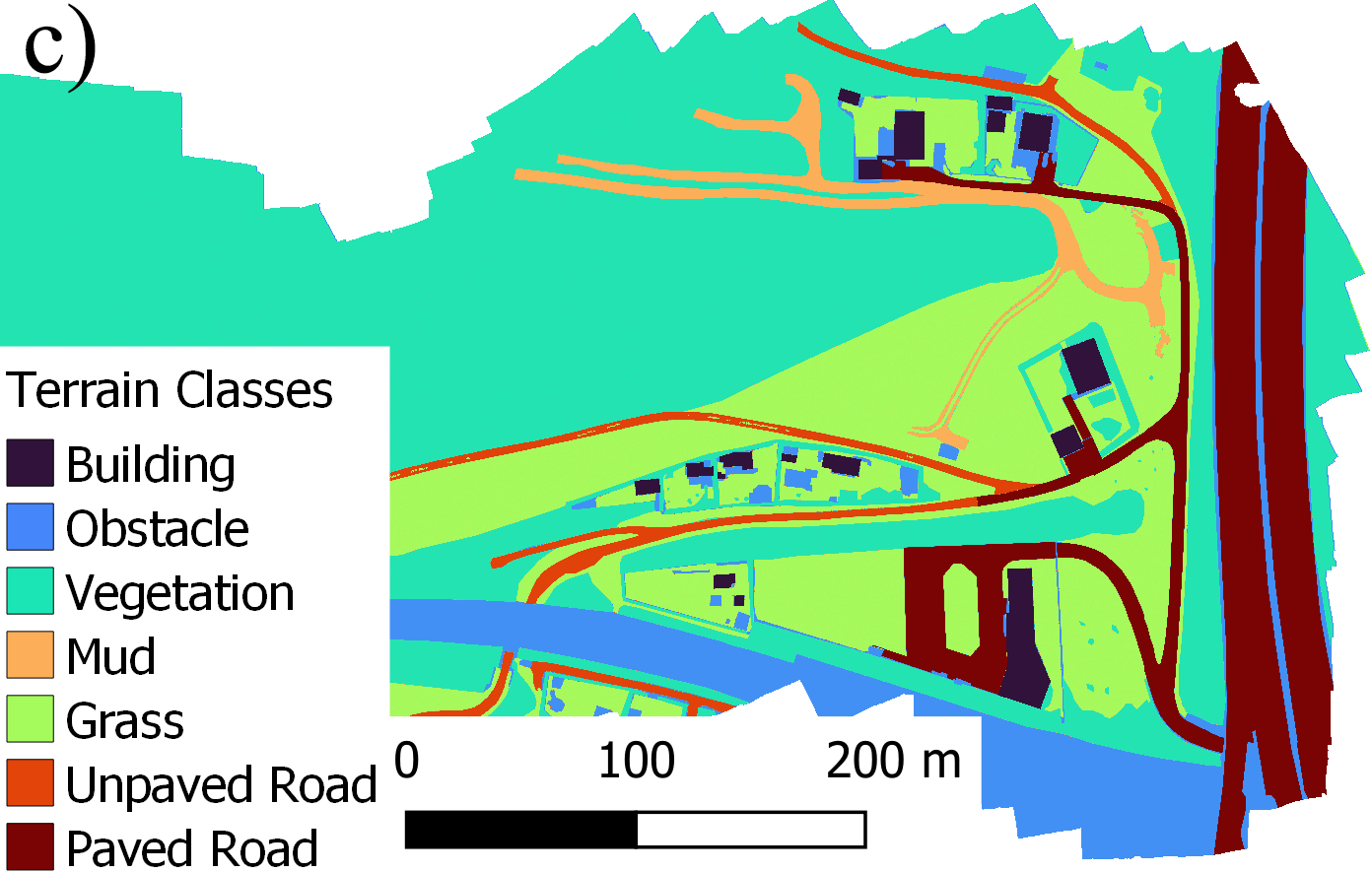}
	\caption{Overview of the test environment. a) orthographic overview of the experimental environment, visualizing the driven path with the robot for data recording. The red area was used for training/testing and the rest for validation, b) the height map of the environment, c) the segmented environment with seven terrain classes of which the latter four are traversable by the robot. White areas do not hold information about the environment.}
	\label{fig:environment}
\end{figure*}

\subsection{Data Collection Process} \label{s-collection}
This section describes the process of building a representation of the environment and the data collection process using a ground robot to record the ground truth data for prediction. 

To build the maps of the environment, a commercial UAV equipped with a high-resolution camera was flown over the recording scene to take overlapping pictures. Photogrammetric processing was then conducted on the recorded images as proposed in \cite{Mayer.2018} using Agisoft Metashape\footnote{https://www.agisoft.com/ (Accessed on Feb 22$^{\text{nd}}$ 2023)} to generate an orthophoto $O$ and 3D spatial data, which was used to generate the height map $H$. $H$ holds altitude information ranging from $574.8m$ to $661.3m$ above sea level  between the lowest and highest recorded point. The orthophoto was manually segmented to annotate the terrain types in $C$. Following the label definitions as used in \cite{Valada.2016}, a total of seven terrain classes were used for annotation, of which four (grass, mud, paved/unpaved road) were set to be traversable by the robot. A visualization of the three environment maps is shown in Figure~\ref{fig:environment}.  The resulting maps covered an area of around $596 m \times 305 m$ and have a resolution of $0.05m$. 

\begin{figure}[!t]
	\centering
        \includegraphics[width=0.49\linewidth]{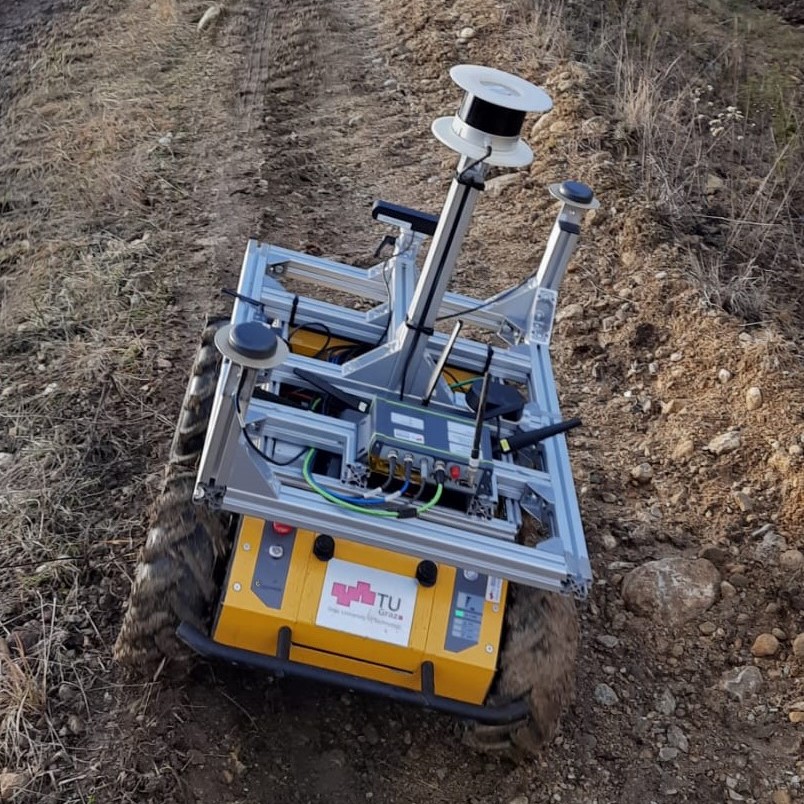}
	\includegraphics[width=0.49\linewidth]{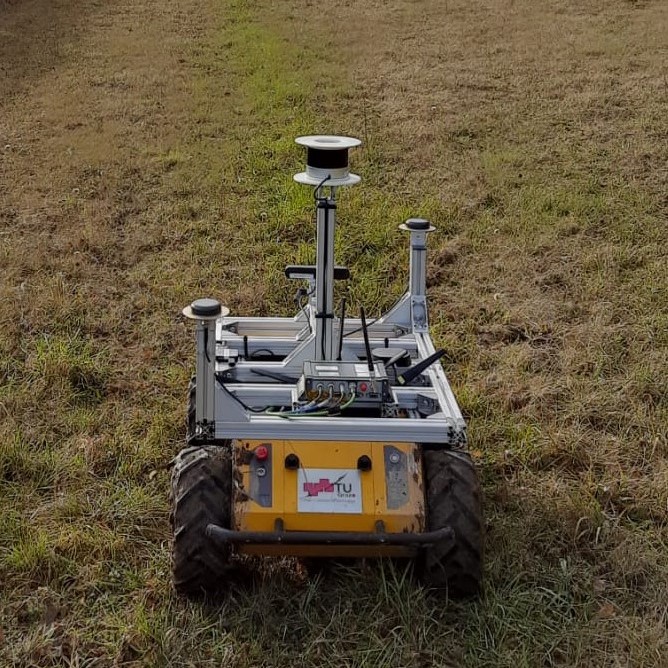}
	\caption{The ground robot used for data collection on two different types of terrain, namely mud (left) and grass (right).}
	\label{fig:hardware}
\end{figure}

To generate training data and ground truth measurements, the differential drive robot Husky from Clearpath Robotics\footnote{https://clearpathrobotics.com/husky-unmanned-ground-vehicle-robot/ (Accessed on Feb 22$^{\text{nd}}$ 2023)} was used. Figure~\ref{fig:hardware} shows the robot in the recording environment on two different types of terrain. It is powered by a lead-acid battery with a nominal voltage of 24V. The system itself can read the battery voltage and current consumed by the motors for propulsion. Moreover, the robot was equipped with a RTK GNSS from GeoKonzept to record the robot's 3D pose with an accuracy of $5cm$. The global positioning data was fused with the robot's wheel odometry, an Xsens MTi-G-710 IMU, and the visual odometry received from a Stereolabs ZED2 using an Unscented Kalman Filter (UKF) to improve the velocity measurements during data recording. For data recording, the robot was guided manually through the environment to cover all the traversable terrains with varying slopes. Figure~\ref{fig:environment}a shows the driven path through the environment which was used for training and validation. During the motion of the robot, the pose, battery voltage, current and fused velocity were recorded by the system at a frequency of $20Hz$. 
To investigate the velocity limits in the environment, the robot was permanently controlled through the environment at its maximum speed of $1.0m/s$. Accelerations and in-place rotations were discarded during the training data generation process.
The slope angles of the robot's recorded path range from $-22.5^{\circ}$ to $22.5^{\circ}$.
In this work, only traversable surfaces are considered for the proposed method. Therefore, we do not consider scenarios where the robot gets stuck which would result in infinite energy consumption and a velocity equal to zero. 
Following the patch extraction process from Section~\ref{s-patch}, a total of $10.886$ patches were extracted of which $51\%$ were on grass, $9\%$ on mud, and $20\%$ on paved and unpaved roads each.

\subsection{Training Sample Generation} \label{s-patch}

\begin{figure}[!t]
	\centering
	\includegraphics[width=\linewidth]{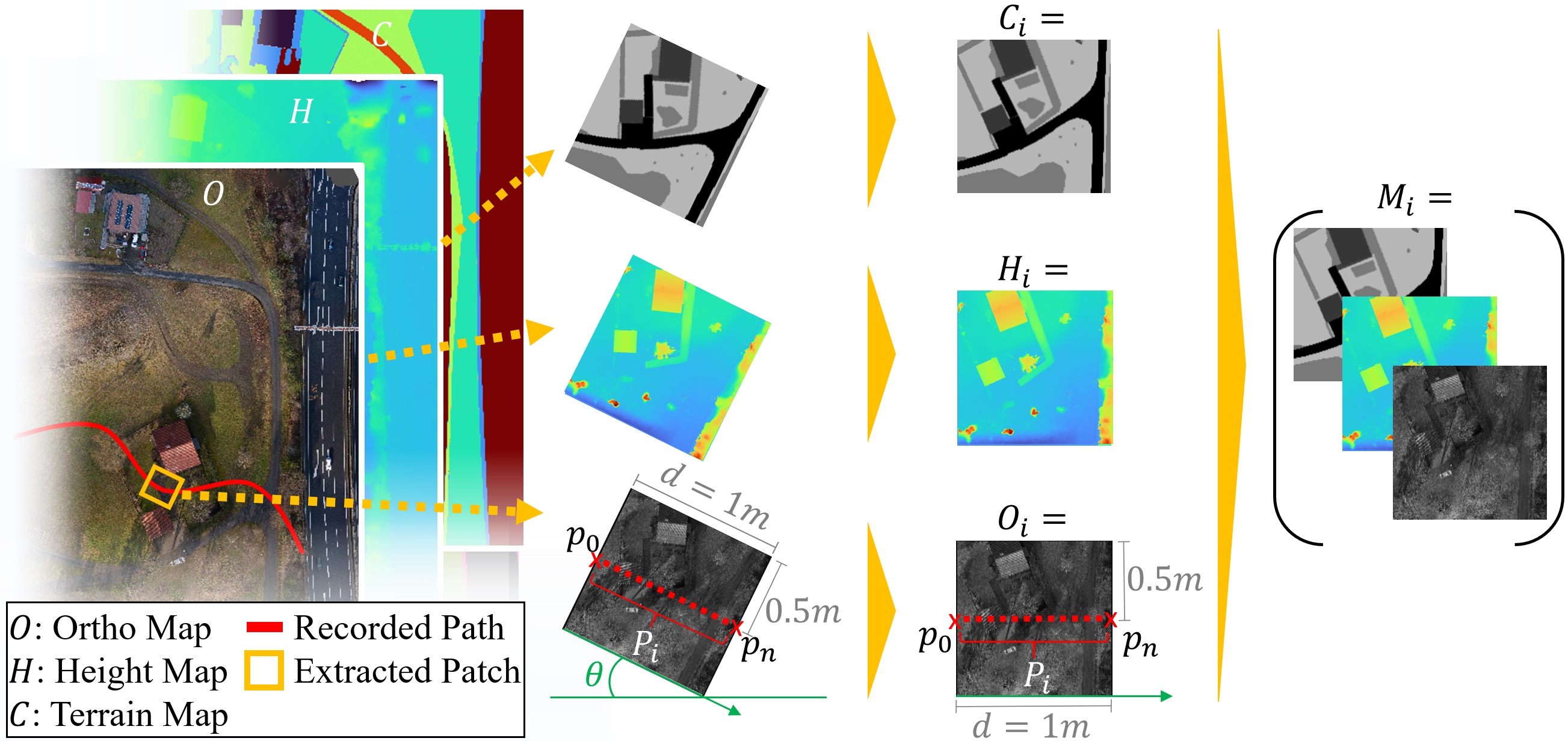}
	\caption{Patch generation process. A patch from each environment map is extracted along the path segment of the robot and rotated to face East. This is used as input for the neural network to predict energy consumption and traversal time for the given patch. }
	\label{fig:rotation}
\end{figure}
To generate training samples for the training function $\hat{f}(M_i)$, we need to extract the environment patches $M_i$. The patch representation is based on the approach from \cite{Wei.2022} and is used as it has information about the path segment $P_i$ encoded and also contains information about the immediate surrounding of the robot. For the training process, we define the length of the path segment to be $d = 1m$ and subsequently extract a patch of size $1m \times 1m$ from all three maps $O,H,C$ held by $\mathcal{M}$. 
Therefore, for any given path segment $P_i = \{p^i_0,...,p^i_n\}$, we place $p^i_0$ on the middle point ($d/2$) on one edge of patch $M_i$ and select $p^i_n$ as the middle point on the opposite edge. This way, the patch is rotated along the robot's path in the environment by $\theta^{\circ}$, where $0^{\circ}$ is heading towards the East. To use the patch as a training sample on existing network architectures, it is rotated by $-\theta^{\circ}$ such that $M_i$ can be represented as a matrix  of size $s\times s \times 3$, with $s=d/r$ depending on the path length $d$ and the resolution $r$ of the environment information $\mathcal{M}$. 
As described in Section~\ref{s-collection}, a resolution $r=0.05m$ is used, resulting in $s=40$ data points per row and column for each layer and thus in a patch size of $M_i = (40 \times 40 \times 3)$. A constant depth of $3$ is chosen for $M_i$ to represent each map ($O,H,C$) in a separate layer. The first layer in $M_i$ holds the grey-scale values origin from the orthophoto $O$, scaled into $[0,1]$. The second layer holds information on the terrain class $C$ which is scaled from $[1,c]$ to $[0,1]$, where $c$ is the number of terrain classes. As described in Section~\ref{s-collection}, a total of $c=7$ terrain classes were used within this work, of which four are traversable by the robot. The third layer holds the relative height information $H$ of the path segment. Therefore, each absolute height value in $M_i$ is subtracted by the minimum height value $\min(H_i)$ of the patch. The height layer is then also scaled into $[0,1]$ using a maximum height difference of $1m$. The height difference was chosen as it represents a $100\%$ grade on the path segment.
 An exemplary visualization of the sampling process is shown in Figure~\ref{fig:rotation}.


\subsection{Network Architecture and Training} \label{s-training}

\begin{figure}[!t]
	\centering
	\includegraphics[width=\linewidth]{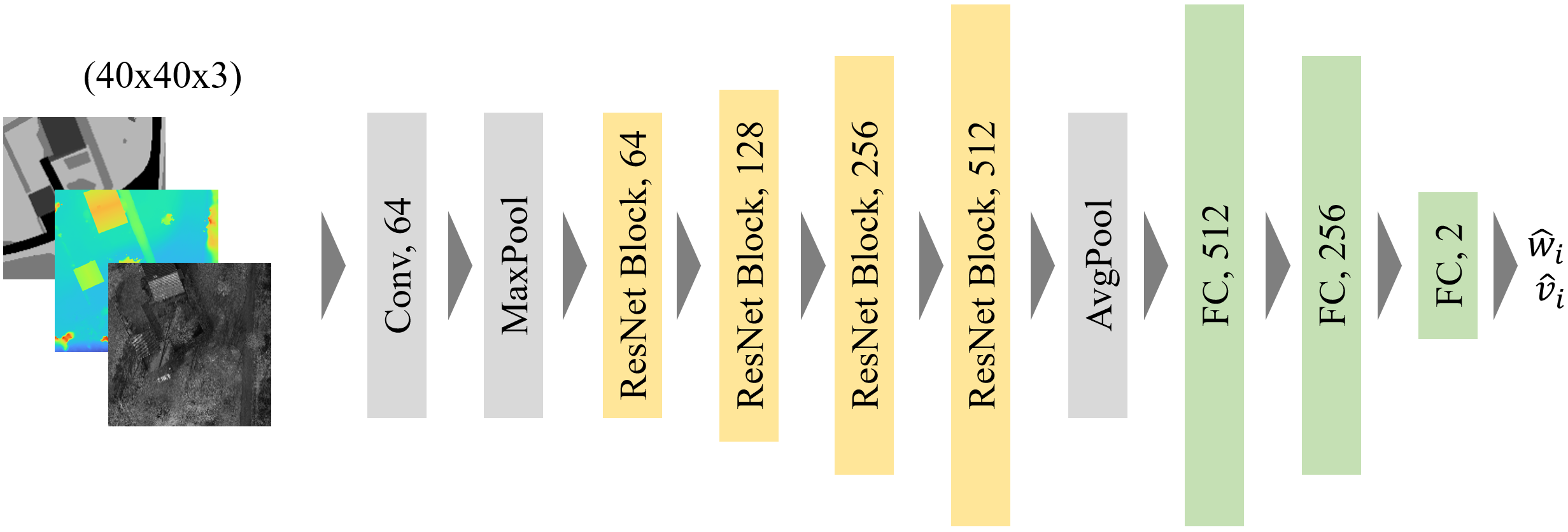}
	\caption{The ResNet18-based network architecture used for training. }
	\label{fig:network}
\end{figure}
To learn the $\hat{f}(M_i)$, the common  ResNet-18 network architecture \cite{7780459} was used with an adapted head for image regression on two variables (see Fig.~\ref{fig:network}), as it has already proven to work for similar tasks \cite{Wei.2022}. 
For the training process, environment patches $M_i$ of size $40\times 40 \times 3$  are used as input and two floating point values $w^*_i,v^*_i$ are used to learn the regression function. As loss function, the NRMSE as described in Equation~\ref{e-loss} was used for the final layer. For training, a learning rate of $10^{-4}$ was used.  The network was trained using Keras.
The collected data from Section~\ref{s-collection} was split into two areas of which one was used for training/testing and one for validation regarding an unseen environment. Figure~\ref{fig:environment}a shows how the collected data was separated. Furthermore, the data for training and testing was split randomly to contain $80\%$ training samples and $20\%$ test samples. This results in a total of $5.142$ training samples, $1.286$ test samples, and $4.458$ samples for validation. The predicted outputs $\hat{w}_i,\hat{v}_i$ of the network were  used to compute the estimated traversal time $T$ and the energy consumption $E$ as shown in Equation~\ref{e-energy}.

\section{Performance Evaluation} \label{s-eval}
In this section, we present the results of our proposed method in different scenarios. First, we present the results of the training environment and show that with the proposed approach it is possible to  predict the traversal time and energy consumption of a given path segment reliably on different types of terrain. Second, we assess the performance of the network in a previously unseen area and confirm the applicability of the trained network in unseen environments. Third, to investigate the effect of the environment information on the anticipated output, we conduct an ablation study on the various input layers. Finally, we demonstrate that our approach outperforms the baseline and eliminates the need to retrain the network for every type of terrain by comparing it to the baseline, presented in \cite{Wei.2022}.

To evaluate the performance of the trained networks, the Root Mean Squared Error (RMSE) and Mean Absolute Percentage Error (MAPE) were chosen as metrics. Both metrics were used for all subsequent evaluations, including the comparison to the baseline. To compute the relative error for a single prediction value, its prediction value ($\hat{x}$) and the ground truth ($x^*$) are used to compute the absolute percentage error (APE): $e = \text{abs}(\hat{x} - x^*)/x^*$.

\subsection{Results on the Known Test Environment} \label{s-test}
To evaluate the applicability of the proposed method on the task of predicting the average power consumption and velocity of the robot driving through a given input patch, we investigated the performance of the trained network on test data that was recorded in the same environment where the data for training the network was recorded.
We summarized the results in Table~\ref{tab:test}, which describes the overall network performance over all given types of terrain ('All') and additionally investigates the performance on each type of terrain to see how well the predictions work on different terrain. While the first two rows show the performance of the trained values $w$ and $v$, the last two rows show the performance of the computed traversal time $T=d/v$ and the energy consumption $E$ as described in Equation~\ref{e-energy}. 

\begin{table}[!t]
\centering
\caption{Performance Results of the proposed approach on the test data, extracted from the same environment which was used for training.}
\resizebox{\columnwidth}{!}{%
\begin{tabular}{
>{\columncolor[HTML]{9B9B9B}}c |
>{\columncolor[HTML]{9B9B9B}}l |ccccc}
\textbf{} & \multicolumn{1}{c|}{\cellcolor[HTML]{9B9B9B}\textbf{}} & \multicolumn{5}{c}{\cellcolor[HTML]{9B9B9B}\textbf{Type of Terrain}} \\
\textbf{Variable} & \multicolumn{1}{c|}{\cellcolor[HTML]{9B9B9B}\textbf{Error}} & \cellcolor[HTML]{9B9B9B}\textbf{Grass} & \cellcolor[HTML]{9B9B9B}\textbf{Mud} & \cellcolor[HTML]{9B9B9B}\textbf{\begin{tabular}[c]{@{}c@{}}Unpaved \\ Road\end{tabular}} & \cellcolor[HTML]{9B9B9B}\textbf{\begin{tabular}[c]{@{}c@{}}Paved\\ Road\end{tabular}} & \cellcolor[HTML]{9B9B9B}\textbf{All} \\ \hline
\cellcolor[HTML]{9B9B9B} & \textbf{RMSE ($W$)} & 14.35 & 13.92 & 7.81 & 16.09 & 14.04 \\
\multirow{-2}{*}{\cellcolor[HTML]{9B9B9B}\textbf{$w$}} & \textbf{MAPE ($\%$)} & 9.96 & 7.51 & 5.74 & 11.53 & 9.39 \\ \hline
\cellcolor[HTML]{9B9B9B} & \textbf{RMSE ($m/s$)} & 0.04 & 0.05 & 0.02 & 0.05 & 0.04 \\
\multirow{-2}{*}{\cellcolor[HTML]{9B9B9B}\textbf{$v$}} & \textbf{MAPE ($\%$)} & 3.32 & 5.27 & 1.78 & 3.64 & 3.41 \\ \hline
\cellcolor[HTML]{9B9B9B} & \textbf{RMSE ($s$)} & 0.06 & 0.10 & 0.03 & 0.07 & 0.06 \\
\multirow{-2}{*}{\cellcolor[HTML]{9B9B9B}\textbf{$T$}} & \textbf{MAPE ($\%$)} & 3.26 & 5.50 & 1.74 & 3.52 & 3.37 \\ \hline
\cellcolor[HTML]{9B9B9B} & \cellcolor[HTML]{9B9B9B}\textbf{RMSE ($J$)} & 18.22 & 29.26 & 10.98 & 25.55 & 21.51 \\
\multirow{-2}{*}{\cellcolor[HTML]{9B9B9B}\textbf{$E$}} & \cellcolor[HTML]{9B9B9B}\textbf{MAPE ($\%$)} & 11.06 & 10.64 & 6.15 & 12.90 & 10.77
\end{tabular}%
}
\label{tab:test}
\end{table}

\begin{figure}[!t]
\centering
\resizebox{\columnwidth}{!}{
\begin{tikzpicture}
  \begin{axis}[
    xmin=1, xmax=84,
    ymin=0, ymax=600,
    ytick distance = 100,
    xlabel={\scriptsize{Patch number}},
    ylabel={\scriptsize{Energy consumption (in $J$)}},
    height=3.9cm,
    x tick label style  ={font=\scriptsize},
    y tick label style  ={/pgf/number format/fixed,/pgf/number format/precision=0,font=\scriptsize},
   legend style={at={(0,1)},anchor=north west},
    grid style=dashed,
    legend columns=2,
    width=0.45\textwidth,
    ylabel shift = -3 pt,
  ]
    \begin{scope}
    \fill[gray,opacity=0.5] ({rel axis cs:0,0}) rectangle ({rel axis cs:0.25,1});
    \fill[gray,opacity=0.5] ({rel axis cs:0.5,0}) rectangle ({rel axis cs:0.75,1});
    \end{scope}
    
    \node[] at (axis cs: 10,400) {\footnotesize Grass};
    \node[text width=1cm,align=center] at (axis cs: 31,350) {\footnotesize Unpaved\\Road};
    \node[text width=1cm,align=center] at (axis cs: 52,350) {\footnotesize Paved\\Road};
    \node[] at (axis cs: 72,150) {\footnotesize Mud};
  \addplot[color=red, error bars/.cd, y dir=both, y explicit] table [col sep=semicolon, x=no, y=joule]{data/test_all_filtered.csv};
  \addplot[color=blue, error bars/.cd, y dir=both, y explicit] table [col sep=semicolon, x=no, y=joule_comp_pred]{data/test_all_filtered.csv};

  \addlegendentry{\scriptsize{Ground Truth $E^*$}}
  \addlegendentry{\scriptsize{Prediction $\hat{E}$}}
  \end{axis}
\end{tikzpicture}}

\caption{Exemplary comparison of the Energy prediction $\hat{E}$ and ground truth $E^*$ of data samples from the test set on different types of terrain.}
\label{fig:test_plot}
\end{figure}
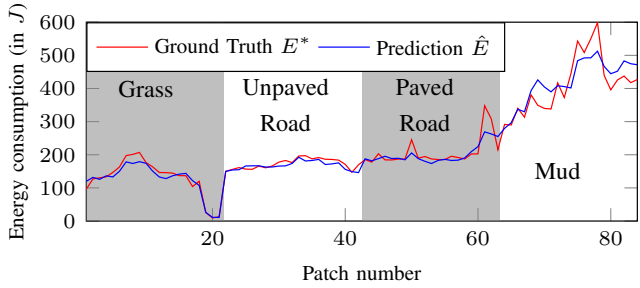

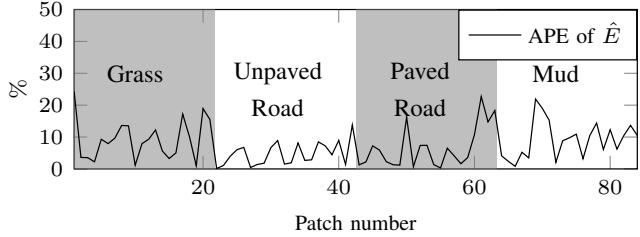
\begin{figure}[!t]
\centering
\resizebox{\columnwidth}{!}{
\begin{tikzpicture}
  \begin{axis}[
    xmin=1, xmax=84,
    ymin=0, ymax=0.5,
    ytick distance = 0.1,
    xlabel={\scriptsize{Patch number}},
    ylabel={\scriptsize{$\%$}},
    height=3.4cm,
    x tick label style  ={font=\scriptsize},
    y tick label style  ={/pgf/number format/fixed,/pgf/number format/precision=2,font=\scriptsize},
    yticklabel={\pgfmathparse{\tick*100}\pgfmathprintnumber{\pgfmathresult}},
    point meta={y*100},
   legend style={at={(1,1)},anchor=north east},
    grid style=dashed,
    legend columns=1,
    width=0.45\textwidth,
    ylabel shift = -3 pt,
  ]
    \begin{scope}
    \fill[gray,opacity=0.5] ({rel axis cs:0,0}) rectangle ({rel axis cs:0.25,1});
    \fill[gray,opacity=0.5] ({rel axis cs:0.5,0}) rectangle ({rel axis cs:0.75,1});
    \end{scope}
    \node[] at (axis cs: 10,.3) {\footnotesize Grass};
    \node[text width=1cm,align=center] at (axis cs: 31,.25) {\footnotesize Unpaved\\Road};
    \node[text width=1cm,align=center] at (axis cs: 52,.25) {\footnotesize Paved\\Road};
     \node[] at (axis cs: 72,.3) {\footnotesize Mud};

  \addplot[color=black, error bars/.cd, y dir=both, y explicit] table [col sep=semicolon, x=no, y=joule_comp_error]{data/test_all_filtered.csv};

  \addlegendentry{\scriptsize{APE of $\hat{E}$}}
  \end{axis}
\end{tikzpicture}}

\caption{Absolute Percentage Error (APE) of the energy prediction $\hat{E}$ for exemplary data samples from the test set on different types of terrain. }
\label{fig:test_error}
\end{figure}

The average error (MAPE) for predicting the power $w$ over all types of terrain is $9.39\%$, and $3.41\%$ for velocity $v$. Using both variables to compute the traversal time and energy consumption leads to an error of $3.37\%$ and $10.77\%$. It can be seen, that the network is trained well on all types of terrain, as the performance on all  terrain types shows a good prediction. The lowest MAPE for $w$ and $v$ could be observed on unpaved roads ($5.74\%$ and $1.78\%$), while the highest MAPE for $w$ was observed on paved roads and for $v$ on mud. 
Overall, a good prediction performance could be observed in the training environment. Figure~\ref{fig:test_plot} shows the prediction performance for energy consumption $E$ on a number of selected data samples from the test dataset on different types of terrain. It can be seen, that the prediction works well for any type of terrain. This is also confirmed when inspecting the relative error $e$ for the same data samples in Figure~\ref{fig:test_error}. 


\subsection{Generalization to unseen Environments} \label{s-genralization}
To show that our trained network generalizes well on unseen environments, we investigate the performance of traversal time $T$ and energy consumption $E$ on the validation dataset, which was selected as shown in Figure~\ref{fig:environment}a. Therefore, a total of $4.458$ data points were used to evaluate the performance of our approach in said environment. Table~\ref{tab:val} shows the results for the predicted values $w,v$, as well as for the computed objective values $T,E$. 

\begin{table}[!t]
\centering
\caption{Performance Results of the proposed approach on the validation data, which was recorded in an environment that was not used during training.}
\resizebox{\columnwidth}{!}{%
\begin{tabular}{
>{\columncolor[HTML]{9B9B9B}}c |
>{\columncolor[HTML]{9B9B9B}}l |ccccc}
\textbf{} & \multicolumn{1}{c|}{\cellcolor[HTML]{9B9B9B}\textbf{}} & \multicolumn{5}{c}{\cellcolor[HTML]{9B9B9B}\textbf{Type of Terrain}} \\
\textbf{Variable} & \multicolumn{1}{c|}{\cellcolor[HTML]{9B9B9B}\textbf{Error}} & \cellcolor[HTML]{9B9B9B}\textbf{Grass} & \cellcolor[HTML]{9B9B9B}\textbf{Mud} & \cellcolor[HTML]{9B9B9B}\textbf{\begin{tabular}[c]{@{}c@{}}Unpaved \\ Road\end{tabular}} & \cellcolor[HTML]{9B9B9B}\textbf{\begin{tabular}[c]{@{}c@{}}Paved\\ Road\end{tabular}} & \cellcolor[HTML]{9B9B9B}\textbf{All} \\ \hline
\cellcolor[HTML]{9B9B9B} & \textbf{RMSE ($W$)} & 18.85 & 29.20 & 15.07 & 16.15 & 17.73 \\
\multirow{-2}{*}{\cellcolor[HTML]{9B9B9B}\textbf{$w$}} & \textbf{MAPE ($\%$)} & 11.29 & 19.25 & 8.42 & 11.58 & 10.74 \\ \hline
\cellcolor[HTML]{9B9B9B} & \textbf{RMSE ($m/s$)} & 0.07 & 0.06 & 0.04 & 0.07 & 0.06 \\
\multirow{-2}{*}{\cellcolor[HTML]{9B9B9B}\textbf{$v$}} & \textbf{MAPE ($\%$)} & 5.60 & 5.22 & 2.90 & 5.70 & 5.14 \\ \hline
\cellcolor[HTML]{9B9B9B} & \textbf{RMSE ($s$)} & 0.09 & 0.09 & 0.04 & 0.09 & 0.08 \\
\multirow{-2}{*}{\cellcolor[HTML]{9B9B9B}\textbf{$T$}} & \textbf{MAPE ($\%$)} & 5.74 & 5.18 & 2.91 & 5.67 & 5.04 \\ \hline
\cellcolor[HTML]{9B9B9B} & \cellcolor[HTML]{9B9B9B}\textbf{RMSE ($J$)} & 27.50 & 49.58 & 19.24 & 22.69 & 26.63 \\
\multirow{-2}{*}{\cellcolor[HTML]{9B9B9B}\textbf{$E$}} & \cellcolor[HTML]{9B9B9B}\textbf{MAPE ($\%$)} & 13.04 & 22.46 & 10.12 & 13.69 & 12.80
\end{tabular}%
}
\label{tab:val}
\end{table}

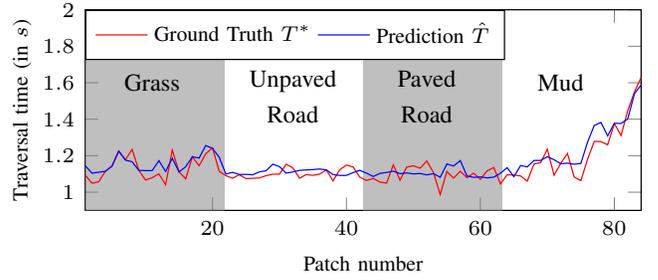
\begin{figure}[!t]
\centering
\resizebox{\columnwidth}{!}{
\begin{tikzpicture}
  \begin{axis}[
    xmin=1, xmax=84,
    ymin=0.9, ymax=2,
    ytick distance = 0.2,
    xlabel={\scriptsize{Patch number}},
    ylabel={\scriptsize{Traversal time (in $s$)}},
    height=3.9cm,
    x tick label style  ={font=\scriptsize},
    y tick label style  ={/pgf/number format/fixed,/pgf/number format/precision=2,font=\scriptsize},
   legend style={at={(0,1)},anchor=north west},
    grid style=dashed,
    legend columns=2,
    width=0.45\textwidth,
    ylabel shift = -3 pt,
  ]
    \begin{scope}
    \fill[gray,opacity=0.5] ({rel axis cs:0,0}) rectangle ({rel axis cs:0.25,1});
    \fill[gray,opacity=0.5] ({rel axis cs:0.5,0}) rectangle ({rel axis cs:0.75,1});
    \end{scope}
    
    \node[] at (axis cs: 11,1.6) {\footnotesize Grass};
    \node[text width=1cm,align=center] at (axis cs: 32,1.525) {\footnotesize Unpaved\\Road};
    \node[text width=1cm,align=center] at (axis cs: 52,1.525) {\footnotesize Paved\\Road};
    \node[] at (axis cs: 72,1.6) {\footnotesize Mud};
  \addplot[color=red, error bars/.cd, y dir=both, y explicit] table [col sep=semicolon, x=no, y=t]{data/d100m25_env2_all_filtered.csv}; 
  \addplot[color=blue, error bars/.cd, y dir=both, y explicit] table [col sep=semicolon, x=no, y=t_comp]{data/d100m25_env2_all_filtered.csv}; 

  \addlegendentry{\scriptsize{Ground Truth $T^*$}}
  \addlegendentry{\scriptsize{Prediction $\hat{T}$}}
  \end{axis}
\end{tikzpicture}}

\caption{Exemplary comparison of the estimated traversal time $\hat{T}$ and ground truth $T^*$ of data samples from the validation set on different types of terrain.}
\label{fig:val_plot}
\end{figure}

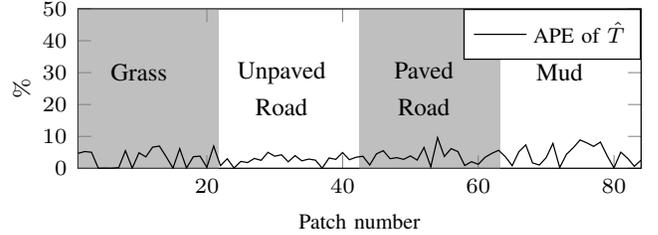
\begin{figure}[!t]
\centering
\resizebox{\columnwidth}{!}{
\begin{tikzpicture}
  \begin{axis}[
    xmin=1, xmax=84,
    ymin=0, ymax=0.5,
    ytick distance = 0.1,
    xlabel={\scriptsize{Patch number}},
    ylabel={\scriptsize{$\%$}},
    height=3.4cm,
    x tick label style  ={font=\scriptsize},
    y tick label style  ={/pgf/number format/fixed,/pgf/number format/precision=2,font=\scriptsize},
    yticklabel={\pgfmathparse{\tick*100}\pgfmathprintnumber{\pgfmathresult}},
    point meta={y*100},
   legend style={at={(1,1)},anchor=north east},
    grid style=dashed,
    legend columns=1,
    width=0.45\textwidth,
    ylabel shift = -3 pt,
  ]
    \begin{scope}
    \fill[gray,opacity=0.5] ({rel axis cs:0,0}) rectangle ({rel axis cs:0.25,1});
    \fill[gray,opacity=0.5] ({rel axis cs:0.5,0}) rectangle ({rel axis cs:0.75,1});
    \end{scope}
    
    \node[] at (axis cs: 10,.3) {\footnotesize Grass};
    \node[text width=1cm,align=center] at (axis cs: 31,.25) {\footnotesize Unpaved\\Road};
    \node[text width=1cm,align=center] at (axis cs: 52,.25) {\footnotesize Paved\\Road};
    \node[] at (axis cs: 72,.3) {\footnotesize Mud};
  \addplot[color=black, error bars/.cd, y dir=both, y explicit] table [col sep=semicolon, x=no, y=t_mape]{data/d100m25_env2_all_filtered.csv};

  \addlegendentry{\scriptsize{APE of $\hat{T}$}}
  \end{axis}
\end{tikzpicture}}

\caption{Absolute Percentage Error (APE) of the estimated traversal time $\hat{T}$ for exemplary data samples from the validation set on different terrain. }
\label{fig:val_error}
\end{figure}

The predictions of the average power $w$ and velocity $v$ on all terrains deviate from the ground truth with a RMSE of $17.73W$ and $0.06m/s$ and a MAPE of $10.74\%$ and $5.14\%$ respectively, which is close to the network's performance in the training environment. Moreover, the network seems to generalize well in nearly all types of terrain. Only the error of $w$ (and consequently $E$) on terrain 'Mud' increases significantly on unseen data. One reason for the error increase could be the relatively small proportion of samples for terrain of type 'Mud' in the validation set ($5.6\%$) which does not fully reflect the properties of the terrain. 
Other than that, the network shows good results in the prediction of traversal time and energy consumption. This verifies that the proposed approach generalizes well to new environments in surrounding areas since the errors are similar to the errors in the training environment. 
Fig.~\ref{fig:val_plot} and Fig.~\ref{fig:val_error} show the prediction performance for traversal time $T$ on a number of selected data samples from the validation dataset on different types of terrain.

\subsection{Ablation of the Input Layers}
To investigate the effect of the given environment information on the trained model, an ablation study was conducted which systematically removes input layers and observes the effect on the predicted outcome. 
The ablation of individual layers has already been proven to be useful to investigate the stability of the trained network on missing features and also to reveal causal relationships of the features  \cite{Meyes.2019}.

To evaluate the stability of the proposed model and to investigate the effect of the environment features, we conducted an ablation of the trained network by systematically replacing environment layers from the network input with random noise. This way, the performance of the trained model could be evaluated with different combinations of input layers. Table~\ref{tab:ablation} gives an overview of the performance results when ablating an environment feature. In the context of this work, the ablation was conducted by not removing the feature completely but by replacing the layer using random noise. The ablation was conducted on the test set ('Test'), as well as on the validation set ('Val'). 

\begin{table}[!t]
\centering
\caption{Performance Results of the proposed approach using ablated input layers. }
\resizebox{\columnwidth}{!}{%
\begin{tabular}{cl|cc|cc|cc|cc}
\rowcolor[HTML]{9B9B9B} 
\textbf{} & \multicolumn{1}{c|}{\cellcolor[HTML]{9B9B9B}\textbf{}} & \multicolumn{2}{c|}{\cellcolor[HTML]{9B9B9B}\textbf{$w$}} & \multicolumn{2}{c|}{\cellcolor[HTML]{9B9B9B}\textbf{$v$}} & \multicolumn{2}{c|}{\cellcolor[HTML]{9B9B9B}\textbf{$E$}} & \multicolumn{2}{c}{\cellcolor[HTML]{9B9B9B}\textbf{$T$}} \\
\rowcolor[HTML]{9B9B9B} 
\textbf{\begin{tabular}[c]{@{}c@{}}Input \\ Patch\end{tabular}} & \multicolumn{1}{c|}{\cellcolor[HTML]{9B9B9B}\textbf{Set}} & \textbf{\begin{tabular}[c]{@{}c@{}}RMSE \\ ($W$)\end{tabular}} & \textbf{\begin{tabular}[c]{@{}c@{}}MAPE \\ ($\%$)\end{tabular}} & \textbf{\begin{tabular}[c]{@{}c@{}}RMSE \\ ($m/s$)\end{tabular}} & \textbf{\begin{tabular}[c]{@{}c@{}}MAPE \\ ($\%$)\end{tabular}} & \textbf{\begin{tabular}[c]{@{}c@{}}RMSE \\ ($J$)\end{tabular}} & \textbf{\begin{tabular}[c]{@{}c@{}}MAPE \\ ($\%$)\end{tabular}} & \textbf{\begin{tabular}[c]{@{}c@{}}RMSE \\ ($s$)\end{tabular}} & \textbf{\begin{tabular}[c]{@{}c@{}}MAPE \\ ($\%$)\end{tabular}} \\ \hline
\cellcolor[HTML]{9B9B9B} & \cellcolor[HTML]{9B9B9B}\textbf{Test} & 14.04 & 9.39 & 0.04 & 3.41 & 21.51 & 10.77 & 0.06 & 3.37 \\
\multirow{-2}{*}{\cellcolor[HTML]{9B9B9B}\textbf{$\{O,H,C\}$}} & \cellcolor[HTML]{9B9B9B}\textbf{Val} & 17.73 & 10.74 & 0.06 & 5.14 & 26.63 & 12.8 & 0.08 & 5.04 \\ \hline
\rowcolor[HTML]{F2F2F2} 
\cellcolor[HTML]{9B9B9B} & \cellcolor[HTML]{9B9B9B}\textbf{Test} & 19.33 & 10.94 & 0.05 & 3.84 & 30.35 & 12.29 & 0.08 & 3.71 \\
\rowcolor[HTML]{F2F2F2} 
\multirow{-2}{*}{\cellcolor[HTML]{9B9B9B}\textbf{$\{H,C\}$}} & \cellcolor[HTML]{9B9B9B}\textbf{Val} & 19.00 & 12.49 & 0.06 & 5.05 & 27.59 & 14.53 & 0.08 & 5.18 \\ \hline
\cellcolor[HTML]{9B9B9B} & \cellcolor[HTML]{9B9B9B}\textbf{Test} & 14.78 & 9.86 & 0.04 & 3.62 & 22.17 & 10.92 & 0.06 & 3.63 \\
\multirow{-2}{*}{\cellcolor[HTML]{9B9B9B}\textbf{$\{O,H\}$}} & \cellcolor[HTML]{9B9B9B}\textbf{Val} & 22.39 & 13.82 & 0.07 & 6.25 & 34.95 & 17.38 & 0.10 & 6.54 \\ \hline
\rowcolor[HTML]{F2F2F2} 
\cellcolor[HTML]{9B9B9B} & \cellcolor[HTML]{9B9B9B}\textbf{Test} & 50.63 & 41.02 & 0.07 & 5.69 & 67.59 & 47.84 & 0.09 & 5.64 \\
\rowcolor[HTML]{F2F2F2} 
\multirow{-2}{*}{\cellcolor[HTML]{9B9B9B}\textbf{$\{O,C\}$}} & \cellcolor[HTML]{9B9B9B}\textbf{Val} & 55.63 & 45.07 & 0.08 & 7.22 & 73.75 & 52.20 & 0.11 & 7.57 \\ \hline
\cellcolor[HTML]{9B9B9B} & \cellcolor[HTML]{9B9B9B}\textbf{Test} & 54.52 & 44.63 & 0.09 & 6.49 & 57.34 & 57.17 & 0.11 & 6.41 \\
\multirow{-2}{*}{\cellcolor[HTML]{9B9B9B}\textbf{$\{O\}$}} & \cellcolor[HTML]{9B9B9B}\textbf{Val} & 61.77 & 50.24 & 0.09 & 8.10 & 87.39 & 60.39 & 0.13 & 8.68 \\ \hline
\rowcolor[HTML]{F2F2F2} 
\cellcolor[HTML]{9B9B9B} & \cellcolor[HTML]{9B9B9B}\textbf{Test} & 66.37 & 44.10 & 0.07 & 5.73 & 91.93 & 47.16 & 0.11 & 5.32 \\
\rowcolor[HTML]{F2F2F2} 
\multirow{-2}{*}{\cellcolor[HTML]{9B9B9B}\textbf{$\{C\}$}} & \cellcolor[HTML]{9B9B9B}\textbf{Val} & 61.49 & 37.11 & 0.07 & 5.92 & 79.83 & 40.07 & 0.10 & 5.98 \\ \hline
\cellcolor[HTML]{9B9B9B} & \cellcolor[HTML]{9B9B9B}\textbf{Test} & 19.33 & 11.69 & 0.05 & 3.71 & 28.67 & 13.31 & 0.07 & 3.60 \\
\multirow{-2}{*}{\cellcolor[HTML]{9B9B9B}\textbf{$\{H\}$}} & \cellcolor[HTML]{9B9B9B}\textbf{Val} & 22.52 & 15.17 & 0.07 & 5.53 & 32.09 & 17.36 & 0.09 & 5.64
\end{tabular}%
}
\label{tab:ablation}
\end{table}

First, only a slight performance decrease can be observed  when removing the ortho patch on the input layer and using $\{H,C\}$ for prediction, suggesting that the network is robust against the loss of ortho information. Similarly to that, the evaluation of $\{O,H\}$ shows that the network structure can cope with the loss of terrain information in the training environment to some extent. However, an MAPE increase in energy prediction $E$ of nearly $5\%$ on the validation set suggests that information on terrain is already an important feature for the generalization of the predictions on new environments.  When removing the height information ($\{O,C\}$), a significant loss in performance can be observed, indicating that height information is the crucial factor to predict energy consumption and traversal time. This is further confirmed when looking at the performance results on a single environment feature used as input. $\{O\}$ is not able to provide stable predictions for the energy consumption and traversal time. 
$\{H\}$ as a single input layer can provide relatively precise predictions, confirming the work of \cite{Wei.2022}. However, to gain the full performance an different types of terrain, the use of all three features as input is still required.

\subsection{Comparison to a baseline approach}
To demonstrate the contribution of our proposed approach, we compare it to baseline approaches which are used to predict the energy consumption $E$ and the traversal time $T$. As no approach exists so far that provides both objectives simultaneously, each objective is compared to its own baseline. We show that our approach performs better than the baseline approaches and also generalizes better to different types of terrain and new environments. For the performance comparison, the same datasets as used in Section~\ref{s-test} and \ref{s-genralization} are used for baseline comparison in the training environment and for an unseen environment. 

\begin{table}[!t]
\centering
\caption{Performance comparison to the baseline approach. 'Wei \& Isler \cite{Wei.2022}' is only trained on terrain 'Grass', while 'Wei \& Isler \cite{Wei.2022} Retrained' consists of multiple networks which were separately trained on each type of terrain.}
\resizebox{\columnwidth}{!}{%
\begin{tabular}{cl|cc|cc|cc}
\rowcolor[HTML]{9B9B9B} 
\textbf{} & \multicolumn{1}{c|}{\cellcolor[HTML]{9B9B9B}\textbf{}} & \multicolumn{2}{c|}{\cellcolor[HTML]{9B9B9B}\textbf{Ours}} & \multicolumn{2}{c|}{\cellcolor[HTML]{9B9B9B}\textbf{\begin{tabular}[c]{@{}c@{}}Wei \& Isler\\ \cite{Wei.2022}\end{tabular}}} & \multicolumn{2}{c}{\cellcolor[HTML]{9B9B9B}\textbf{\begin{tabular}[c]{@{}c@{}}Wei \& Isler \cite{Wei.2022}\\ Retrained\end{tabular}}} \\
\rowcolor[HTML]{9B9B9B} 
\textbf{\begin{tabular}[c]{@{}c@{}}Prediction\\ Value\end{tabular}} & \multicolumn{1}{c|}{\cellcolor[HTML]{9B9B9B}\textbf{Set}} & \textbf{\begin{tabular}[c]{@{}c@{}}RMSE \\ ($J$)\end{tabular}} & \textbf{\begin{tabular}[c]{@{}c@{}}MAPE \\ ($\%$)\end{tabular}} & \textbf{\begin{tabular}[c]{@{}c@{}}RMSE \\ ($J$)\end{tabular}} & \textbf{\begin{tabular}[c]{@{}c@{}}MAPE \\ ($\%$)\end{tabular}} & \textbf{\begin{tabular}[c]{@{}c@{}}RMSE \\ ($J$)\end{tabular}} & \textbf{\begin{tabular}[c]{@{}c@{}}MAPE \\ ($\%$)\end{tabular}} \\ \hline
\rowcolor[HTML]{FFFFFF} 
\cellcolor[HTML]{9B9B9B} & \cellcolor[HTML]{9B9B9B}\textbf{Test} & 21.51 & 10.77 & 51.39 & 27.25 & 34.52 & 14.64 \\
\rowcolor[HTML]{FFFFFF} 
\multirow{-2}{*}{\cellcolor[HTML]{9B9B9B}\textbf{$E$}} & \cellcolor[HTML]{9B9B9B}\textbf{Val} & 26.63 & 12.80 & 56.16 & 28.37 & 32.92 & 16.96
\end{tabular}%
}
\label{tab:energy_comparison}
\end{table}

For the comparison of the energy consumption, the method proposed by Wei and Isler \cite{Wei.2022} is used as a baseline, as it has been proven to work well in outdoor terrains and outperforms certain physics-based models. However, it has to be noted that the baseline approach is not able to adapt to new types of terrain within a single trained network. Therefore, our approach is compared to two variants of the baseline: The first variant is the network trained solely on data samples recorded on the terrain of type 'Grass' (as done in \cite{Wei.2022}). For the second variant, the approach from \cite{Wei.2022} is retrained separately on every type of terrain to show the best possible performance of the baseline method on the given terrain class. Table~\ref{tab:energy_comparison} shows the overall performance comparison of our approach to the baseline. As can be seen, when evaluated on multiple types of terrain, our approach outperforms the baseline approach which was trained on grassy terrain by more than $15\%$. 
This proves that our approach can adapt to different types of terrain. Moreover, our approach outperforms the baseline by $5\%$ in both the training and validation environments when compared to the overall performance of the retrained baseline. This demonstrates that the proposed approach can further improve energy prediction in outdoor environments on multiple types of terrain and also generalizes better to unseen environments. 


To investigate the performance of the traversal time estimation, our approach is compared to the expected traversal time which is based on the theoretical system limitations of the robot. We compare the expected traversal time to the outcome of our approach, which takes into account the robot's immediate surroundings while moving through a patch, to demonstrate that the expected traversable time also depends on environmental factors such as slope and type of terrain. Therefore, as a baseline, the robot's expected velocity $v_e$ along a data sample is divided by the length $d$ of a path segment  such that $T_e = d/v_e$. As the robot is driven through the environment at the maximum speed of $v_e=1m/s$ and the distance of a path segment in the recorded dataset is $d=1$, the expected traversal time is a constant factor of one second. Table~\ref{tab:time_comparison} shows how the consideration of environmental limits can improve the traversal time estimation, when compared to the expected traversal time. It can be seen that the standard deviation (RMSE) of the actual time and expected time is within $0.17-0.22s$ causing an error of up to $17.45\%$. By estimating the velocity limit under consideration of the environment, an overall error (MAPE) of $~5\%$ can be achieved. This shows, that the consideration of the robot's limits in different environmental settings is an important factor when planning a time-optimal path through the environment.

\begin{table}[!t]
\centering
\caption{Performance comparison to the expected time. }
\resizebox{0.85\columnwidth}{!}{%
\begin{tabular}{cl|cc|cc}
\rowcolor[HTML]{9B9B9B} 
\textbf{} & \multicolumn{1}{c|}{\cellcolor[HTML]{9B9B9B}\textbf{}} & \multicolumn{2}{c|}{\cellcolor[HTML]{9B9B9B}\textbf{Ours}} & \multicolumn{2}{c}{\cellcolor[HTML]{9B9B9B}\textbf{Expected Time}} \\
\rowcolor[HTML]{9B9B9B} 
\textbf{\begin{tabular}[c]{@{}c@{}}Prediction\\ Value\end{tabular}} & \multicolumn{1}{c|}{\cellcolor[HTML]{9B9B9B}\textbf{Set}} & \textbf{\begin{tabular}[c]{@{}c@{}}RMSE \\ ($s$)\end{tabular}} & \textbf{\begin{tabular}[c]{@{}c@{}}MAPE \\ ($\%$)\end{tabular}} & \textbf{\begin{tabular}[c]{@{}c@{}}RMSE \\ ($s$)\end{tabular}} & \textbf{\begin{tabular}[c]{@{}c@{}}MAPE \\ ($\%$)\end{tabular}} \\ \hline
\rowcolor[HTML]{FFFFFF} 
\cellcolor[HTML]{9B9B9B} & \cellcolor[HTML]{9B9B9B}\textbf{Test} & 0.06 & 3.37 & 0.22 & 17.45 \\
\rowcolor[HTML]{FFFFFF} 
\multirow{-2}{*}{\cellcolor[HTML]{9B9B9B}\textbf{$T$}} & \cellcolor[HTML]{9B9B9B}\textbf{Val} & 0.08 & 5.04 & 0.17 & 13.82
\end{tabular}%
}
\label{tab:time_comparison}
\end{table}

\section{Conclusion} \label{s-conclusion}
In this paper, we presented an approach to predict the energy consumption and traversal time of a ground robot for path planning on different types of terrain in outdoor environments. We showed that in our trained machine-learning model, height information has the biggest impact on predicting the energy consumption and traversal time of a given path. However, to improve the overall prediction performance, more information about the environment and type of terrain has to be considered. Therefore, we showed that the incorporation of information on the type of terrain and visual information from an orthophoto is a valuable factor to improve the prediction performance.  We showed that our method significantly reduces the error when  predicting the energy consumption when compared to a baseline approach  and also gets rid of the necessity to retrain a model for every type of terrain.  
We also showed that the expected traversal time based on simple robot definitions from a fact sheet cannot be used to precisely predict the traversal time in complex terrain. Our approach, however, takes into account a robot's immediate surroundings to improve the estimation for traversal time. 
To further improve the proposed approach, future work can incorporate rotational velocities in the model, as rotations also play a crucial role in accurately predicting the robot's energy consumption and traversal time. Furthermore, the model can be used as a baseline to adapt the trained information to new circumstances, such as a new robot system or changing weather/seasonal conditions.



\bibliographystyle{IEEEtran}
\bibliography{IEEEabrv,IEEEexample}

\end{document}